\documentclass[letterpaper, 10pt, conference]{IEEEtran}
\IEEEoverridecommandlockouts
\usepackage{microtype}
\usepackage{url}
\usepackage[hidelinks]{hyperref}
\usepackage[utf8]{inputenc}
\usepackage[small]{caption}
\usepackage{graphicx}
\usepackage{booktabs}
\usepackage{algorithm}
\usepackage{algorithmic}
\usepackage{array}
\usepackage{verbatim}
\usepackage{amsmath,amssymb,amsfonts}
\usepackage{mathtools}
\usepackage[dvipsnames]{xcolor}
\usepackage{balance}
\usepackage[normalem]{ulem}
\usepackage{comment}
\usepackage{csquotes}
\usepackage{subcaption}
\usepackage{bm}
\usepackage{siunitx}
\usepackage[formats]{listings}
\usepackage[nameinlink]{cleveref}
\usepackage{enumerate}
\usepackage{pifont}

\DeclareMathOperator*{\argmin}{argmin}
\DeclareMathOperator*{\argmax}{argmax}

\floatstyle{plain} %
\newfloat{listing}{htbp}{lop}
\floatname{listing}{Listing}

\crefname{subfigure}{Subfigure}{Subfigures}
\Crefname{subfigure}{Subfigure}{Subfigures}

\begin{document}
\bstctlcite{IEEEexample:BSTcontrol}

\title{
    The Constitutional Controller:
    Doubt-Calibrated Steering of Compliant Agents
}

\author{ 
    Simon Kohaut$^{1}$, Felix Divo$^{1}$, Navid Hamid$^{1}$, Benedict Flade$^{2}$,\\Julian Eggert$^{2}$, Devendra Singh Dhami$^{3}$, Kristian Kersting$^{1,4,5,6}$
    \thanks{
        $^{1}$ Artificial Intelligence and Machine Learning Lab, TU Darmstadt
    }%
    \thanks{
        $^{2}$ Honda Research Institute Europe
    }%
    \thanks{
        $^{3}$ Uncertainty in Artificial Intelligence Group, TU Eindhoven
    }
    \thanks{
        $^{4}$ Hessian AI
    }%
    \thanks{
        $^{5}$ Centre for Cognitive Science
    }%
    \thanks{
        $^{6}$ German Center for Artificial Intelligence (DFKI)
    }%
    \thanks{
        $^{7}$ \url{https://github.com/simon-kohaut/CoCo}
    }%
}

\maketitle

\begin{abstract}
Ensuring reliable and rule-compliant behavior of autonomous agents in uncertain environments remains a fundamental challenge in modern robotics.
Our work shows how neuro-symbolic systems, which integrate probabilistic, symbolic white-box reasoning models with deep learning methods, offer a powerful solution to this challenge.
They enable the simultaneous consideration of explicit rules and neural models trained on noisy data, combining the strengths of structured reasoning with flexible representations.
To this end, we introduce the Constitutional Controller (CoCo), a novel framework designed to enhance the safety and reliability of agents by reasoning over deep probabilistic logic programs that represent constraints, such as those found in shared traffic spaces.
Furthermore, we propose the concept of self-doubt, implemented as a probability density conditioned on doubt features such as travel velocity, employed sensors, or health factors.
In a real-world aerial mobility study, we demonstrate CoCo’s advantages for intelligent autonomous systems to learn appropriate doubts and navigate complex, uncertain environments safely and compliantly.
\end{abstract}

\section{Introduction}

\PARstart{T}{rustworthy} decision-making is essential for deploying autonomous agents in human-inhabited spaces. 
These systems must not only handle environmental uncertainty but also be interpretable and provably compliant with complex regulations~\cite{TIGER2020325}.
This challenge is particularly acute for Unmanned Aerial Vehicles (UAVs) in future Advanced Air Mobility (AAM) applications, ranging from logistics to emergency response~\cite{goyal2022advanced}. 
The key difficulty lies in creating systems that can represent, reason about, and provably adhere to intricate legal and safety constraints, such as no-fly zones, which go far beyond simple crash mitigation~\cite{PURTELL2024102569}.
Neuro-symbolic approaches, which connect logical reasoning with probabilistic inference, offer a powerful and interpretable way to represent these restrictions. 
Building on this foundation, we present the Constitutional Controller (CoCo), a novel framework that directly couples an agent's motion control to a neuro-symbolic model of its operational context ("Constitution"), representing its perception and environment in conjunction with navigational constraints, such as traffic laws or user preferences.
\begin{figure}[t]
    \centering
    \includegraphics[width=\linewidth]{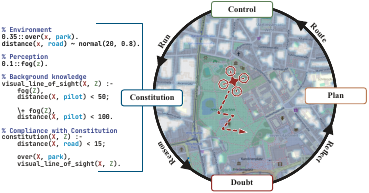}
    \caption{
        \textbf{Constitutional Control for navigating complex, regulated spaces:}
        CoCo integrates neuro-symbolic reasoning and doubt modeling to steer its actions towards safety and compliance.
    }
    \label{fig:motivation}
\end{figure}
While prior work has applied similar models to high-level mission design~\cite{kohaut2023mission} or agent prediction~\cite{kohaut2025constitutional}, CoCo is the first framework to integrate them directly into the \textit{low-level motion control loop}. 
CoCo achieves this by integrating deep probabilistic first-order logic to capture legal and physical constraints, which can be sourced from experts or Large Language Models~\cite{kohaut2025nesyits}. 
Critically, however, to operate responsibly in the control loop, CoCo must account for the agent's own physical limitations and uncertainties.
With CoCo, we therefore propose modeling an agent's "self-doubt" in conjunction with the neuro-symbolic reasoning of the Constitution.
It is represented as a learned, conditional density of the agent's own physical inaccuracies, such as control errors under varying conditions. 
This model of self-doubt enables the agent to plan and act in a mindful, verifiable manner in compliance with its rules.
Hence, our contributions are threefold:
(i) The Constitutional Controller (CoCo), a novel control framework that directly couples low-level motion control with a neuro-symbolic model of operational constraints.
(ii) The formulation of agent doubt, a learned, conditional probabilistic model of an agent's own physical inaccuracies used to ensure compliance.
(iii) A real-world demonstration on a physical UAV validating that CoCo uses learned doubt to adapt its behavior, measurably improving safety and regulatory adherence.
Furthermore, we provide an open-source implementation of CoCo alongside this work${}^7$.

\section{Related Work}

\subsection{Neuro-Symbolic Systems}

While early probabilistic logic languages~\cite{bayesian_logic,problog,inference_in_plp} did not consider end-to-end learning with neural networks, languages such as DeepProbLog~\cite{deepproblog}, NeurASP~\cite{neurasp}, and SLASH~\cite{slash} close this gap and combine the strengths of neural and probabilistic reasoning.
Recent work explored the application of such approaches for Probabilistic Mission Design (ProMis)~\cite{kohaut2023mission,kohaut2024ceo} by encoding, e.g., traffic laws, in hybrid probabilistic first-order logic~\cite{nitti2016probabilistic}.

CoCo's Constitution builds upon hybrid probabilistic logic programming for incorporating spatial relations in an uncertain environment, e.g., as obtained from neural computer vision models or low-cost sensors~\cite{Flade2020}. 
Furthermore, we introduce syntax and semantics for CoCo's doubt quantification within the Constitution, specifying the interface to a neural model that acts as the conditional probability density of an agent's self-doubts about its capabilities.
Hence, in addition to leveraging neuro-symbolic systems as building blocks of CoCo's internal knowledge representation, we expand on both the symbolic and sub-symbolic levels of representation.

\subsection{Environment Representation}
\label{sec:related:env_repr}

Reasoning about an agent's actions requires suitable environment representations that model the mission space, obstacles, and viable paths, along with the underlying uncertainties stemming from sensor noise or neural perception.
Such representations can be discrete, such as Occupancy Grid Maps~\cite{thrun2002probabilistic,jun2003probability}, which have been applied, e.g., in SLAM~\cite{grisetti2005improving}, tracking~\cite{chen2006dynamic}, and path planning~\cite{himmelsbach2008lidar}.
Alternatively, representations may be continuous, modeling the environment as scalar fields, which can be estimated using Gaussian Process (GP) regression to capture field intensities in regions of uncertainty, as shown by Qureshi et al.~\cite{qureshi2024scalar}.
Both types of maps allow the representation of the physical environment while modeling uncertainty in sensing or occupancy.
However, less attention has been paid to representing and enforcing high-level legal and behavioral constraints directly within the environment model.

While some logic-based representations allow for reasoning about temporal rules and agent behavior~\cite{TIGER2020325}, their uncertainty representation focuses on the agent itself rather than the environment.
To bridge this gap, Statistical Relational Maps (StaR Maps) have been proposed~\cite{flade2023star} as a hybrid probabilistic environment representation, employing both categorical and continuous distributions to model spatial relations in noisy maps and interface with neuro-symbolic systems.
In CoCo, we leverage StaR Maps to represent and reason about the agent's environment in a probabilistic manner, expressing, e.g., expert knowledge and regulations.

\subsection{Probabilistic Robotics}
In navigating uncertain environments, Bayesian filters, such as the Kalman Filter family~\cite{kalman1960new, julier1995new, julier1997new}, are often used.
Analogously to our work on Constitutional Control, the Constitutional Filter has been shown to extend Probabilistic Mission Design towards learning and leveraging trust into other agents in shared traffic spaces for Bayesian estimation~\cite{kohaut2025constitutional}.
CoCo, on the other hand, is based on the idea of considering self-doubt in tandem with the concept of an agent's own Constitution for safety and compliance.

Mindful planning is crucial when defining a mission as it involves generating a collision-free path to a target location while considering various cost functions under kinodynamic constraints~\cite{yang2014literature,Probst2021}.
Well-known methods for path planning range from graph-based approaches that search for optimal and shortest path solutions, such as A* \cite{hart1968formal,musliman2008implementing}, to sampling-based methods such as Rapid Random Trees~\cite{yang2008real}, and evolutionary hybrid approaches~\cite{hohmann2021hybrid,hohmann2022multi}.

We follow the line of work on Probabilistic Mission Landscapes as cost landscapes for navigation under uncertainty~\cite{kohaut2024ceo}, expanding on ProMis' probabilistic representation and reasoning by adjusting the areas that satisfy the mission's logical constraints. 
Specifically, we enhance it with agent- and context-dependent self-doubt for quantifying the uncertainty of an agent's control module and localization.

\section{Preliminaries}

\subsection{Dynamical Systems}
Dynamical systems provide a formal framework for representing how a robot's state $\mathbf{x} \in \mathcal{X} \subseteq \mathbb{R}^X$, $X \in \mathbb{N}$, evolves over time under the influence of control inputs $\mathbf{u}_t \in \mathbb{R}^U$, $U \in \mathbb{N}$, as well as environmental disturbances. 
Hereby, the state $\mathbf{x}$ encapsulates all necessary variables to describe the robot's configuration and motion, such as position, velocity, orientation, and joint angles, at a given moment. 

Dynamical systems are commonly modeled such that
\begin{equation}
    \mathbf{x}_t = f(\mathbf{x}_{t-1}, \mathbf{u}_t, \mathbf{e}_{x,t}) 
    \quad \text{and} \quad
    \mathbf{z}_t = h(\mathbf{x}_t, \mathbf{e}_{z,t}),
\end{equation}
where $\mathbf{x}$ is governed by a process model $f$ describing the system dynamics, and $h$ is the observation or output model describing a mapping from states to sensor readings $\mathbf{z}_t \in \mathcal{Z} \subseteq \mathbb{R}^Z$, $Z \in \mathbb{N}$. 
Both are subject to stochastic noise terms $\mathbf{e}_{x,t}$ and $\mathbf{e}_{z,t}$, such as fluctuations in air pressure disturbing a UAV or multipath effects on GPS readings.

\begin{figure}[t]
    \centering
    \begin{subfigure}{0.525\linewidth}
        \includegraphics[width=\textwidth]{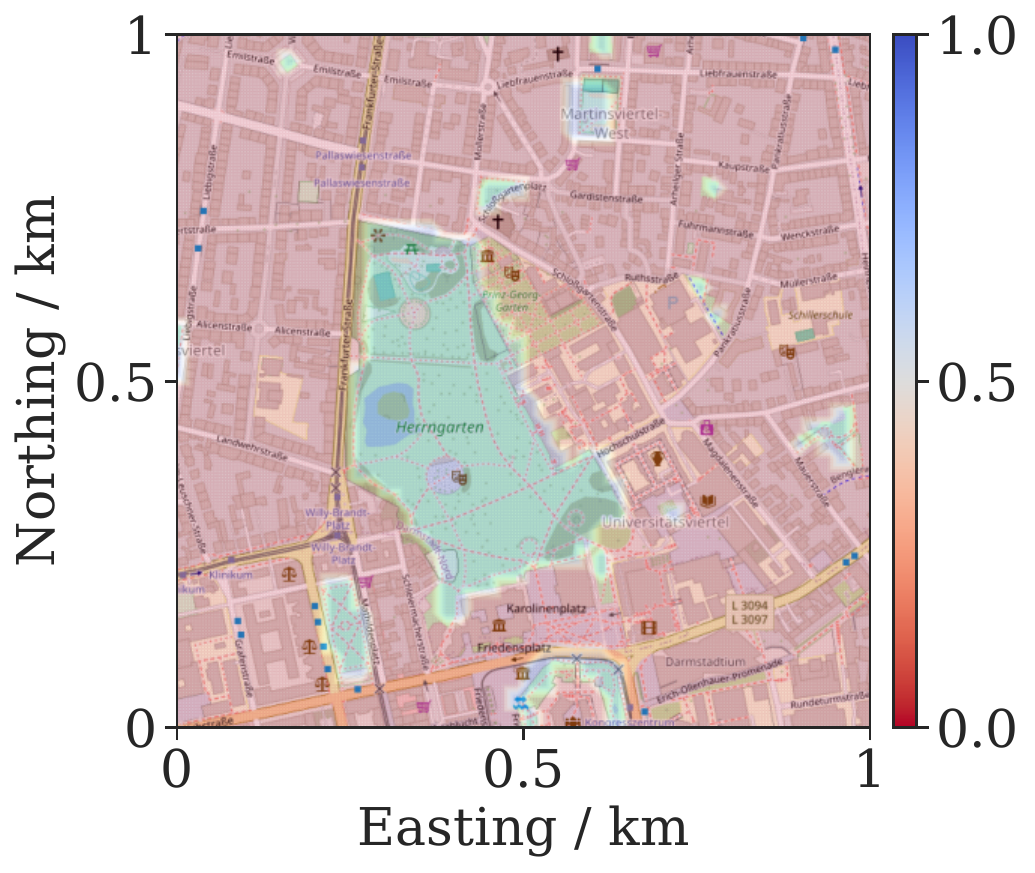}
        \caption{$\mathbb{E}\left[over(X, park)\right]$}
    \end{subfigure} 
    \hfill
    \begin{subfigure}{0.445\linewidth}
        \includegraphics[width=\textwidth]{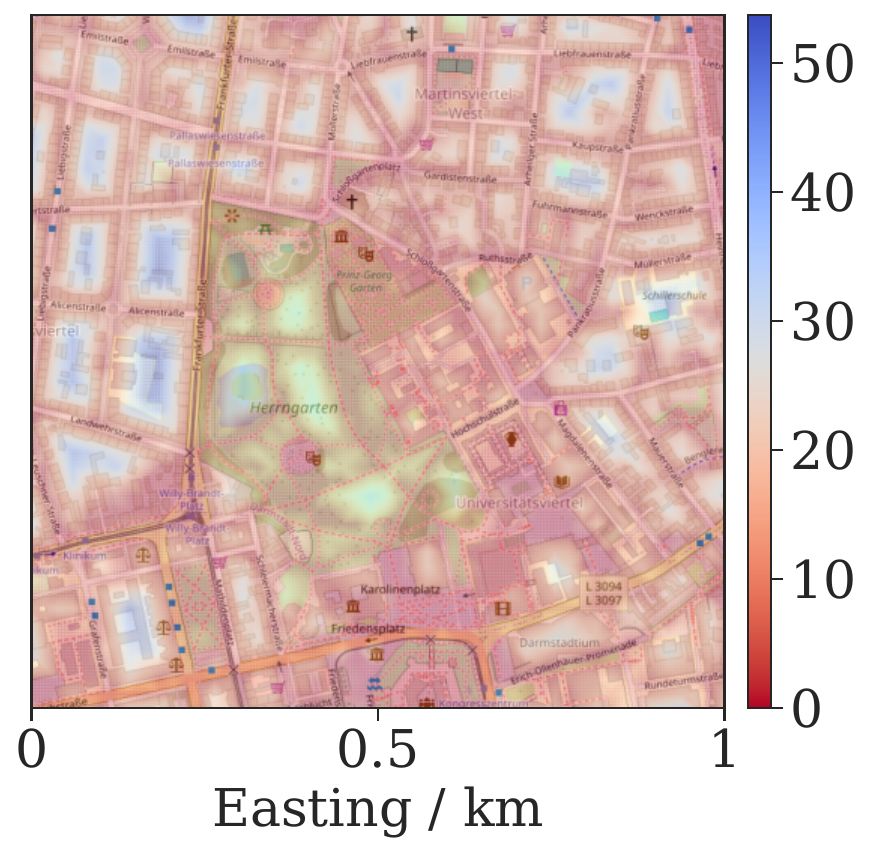}
        \caption{$\mathbb{E}\left[distance(X, road)\right]$}
    \end{subfigure}
    \caption{
        \textbf{StaR Maps in an urban environment}:
        We show the expected values of the \textit{over} and \textit{distance} spatial relations.
    }
    \label{fig:spatial_relations}
\end{figure}

\subsection{Statistical Relational Maps}
\label{sec:starmaps}

Statistical Relational Maps~(StaR Maps) offer a powerful model to capture environments in a probabilistic fashion~\cite{flade2023star}.
Consider the relations \textit{over(x, g)} and \textit{distance(x, g)} shown in \Cref{fig:spatial_relations} between a point $\mathbf{x} \in \mathbb{R}^D$ and environment features $g \in \mathcal{G}$, e.g., $D=2$ and $\mathcal{G} = \{park, road, pilot\}$ in the drone flight scenario from~\Cref{fig:motivation}.
To estimate the parameters of these relations, consider a map $\mathcal{M} = (\mathcal{V}, \mathcal{E}, \rho)$ as a triple of vertices $\mathcal{V}$, edges $\mathcal{E}$ between them, and a tagging function $\rho : \mathcal{V} \to \mathcal{P}(\mathcal{G})$.
The function $\rho(\mathbf{v}) \subseteq \mathcal{G}$ annotates vertex $\mathbf{v} \in \mathcal{V}$ with a set of semantic tags.
If a path exists between two vertices in $\mathcal{V}$ across edges in $\mathcal{E}$, StaR Maps considers them part of the same geometric \textit{feature}.

StaR Maps employ a stochastic error model as established by Flade et al.~\cite{flade2021error} to capture inaccuracies from real-world map data.
That is, for each $\mathbf{v}_{i, j} \in \mathcal{V}$, being the $j$-th vertex of the $i$-th feature, one generates $N \in \mathbb{N}$ randomly affine transformed samples:
\begin{align*}
    \mathbf{\Phi}^{(n)} &\sim \phi_i \tag*{(Linear Map)} \\
    \mathbf{t}^{(n)} &\sim \kappa_i \tag*{(Translation)} \\
    \mathbf{v}^{(n)}_{i,j} &= \mathbf{\Phi}^{(n)} \cdot \mathbf{v}_{i, j} + \mathbf{t}^{(n)} \tag*{(Affine Transformation)}
\end{align*}
Here, $\phi_i$ and $\kappa_i$ are distributions of linear maps $\mathbf{\Phi}^{(n)}$ per feature $i$, such as rotation, scaling, or shearing, and translations $\mathbf{t}^{(n)}$.
By taking the $n$-th sample of each vertex and a copy of the edges $\mathcal{E}$, we obtain a set of $N$ randomly transformed maps $\mathcal{M}^{(n)}$.
This allows computing statistical moments of the spatial relations.
To this end, let $r(\mathcal{M}^{(n)}, \mathbf{x}, g)$ be a deterministic function evaluating a spatial relation on map $\mathcal{M}^{(n)}$ at location $\mathbf{x}$ limited to features for which $\rho(\mathbf{v}) = g$.
One can therefore estimate the parameters of the distribution
\begin{align}
    \widehat{\mu_r} &= \frac{1}{N} \sum_n r(\mathcal{M}^{(n)}, \mathbf{x}, g), \\
    \widehat{\sigma^2_r} &= \frac{1}{N - 1} \sum_n \left(r(\mathcal{M}^{(n)}, \mathbf{x}, g) - \widehat{\mu_r}\right)^2.
\end{align}%
For example, assume \textit{distance} to be a normally distributed random variable with the deterministic function $d$, StaR Maps then yields the density \textit{distance}$(\mathbf{x}, g) \sim \mathcal{N}(\widehat{\mu_d}, \widehat{\sigma^2_d})$.

\section{Method}

\subsection{Architecture}
\begin{figure}[t]
    \centering
    \includegraphics[width=\linewidth]{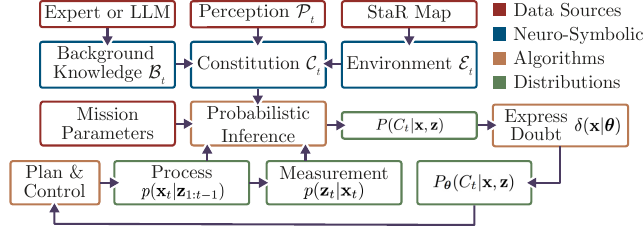}
    \caption{
        \textbf{The Constitutional Controller's architecture:}
        CoCo reasons on a neuro-symbolic model of the task's rules while applying its conditional doubt density on the Constitution's cost landscape.
    }
    \label{fig:architecture}
\end{figure}

The Constitutional Controller (CoCo) is a novel architecture for safe steering of compliant agents in uncertain environments.
As illustrated in \Cref{fig:architecture}, CoCo integrates four major ingredients of knowledge representation:
(i) expert knowledge about environment restrictions, e.g., traffic laws, encoded in first-order logic, (ii) neural and sensor perception, (iii) StaR Maps for spatial reasoning in noisy environments, and (iv) a neural self-doubt model expressing contextualized beliefs in an agent's capabilities during mission execution.

\subsection{Neuro-Symbolic Constitutions}
The \textbf{Constitution} is the agent’s structured knowledge base; a logical model that represents what it knows, what it sees, and how it models its environment to inform its decisions.
We encode the agent's Constitution as an amalgamation of three deep probabilistic first-order logic programs $\mathcal{C}_t = \mathcal{B}_t \cup \mathcal{P}_t \cup \mathcal{E}_t$, combining background knowledge $\mathcal{B}_t$, perception $\mathcal{P}_t$, and environment representation $\mathcal{E}_t$ as illustrated in \Cref{fig:motivation}.
We assume $\mathcal{B}_t$ to be provided by a domain expert, i.e., by formalizing their knowledge as first-order logic or by translating natural language descriptions through a Large Language Model~\cite{rabiei2025LTLCodeGen}.
Furthermore, $\mathcal{P}_t$ encodes the perceived knowledge of the environment obtained from an agent's (neural) sensing capabilities, e.g., information about other traffic participants, weather conditions, or the agent itself.
Finally, we employ a StaR Map to automatically create $\mathcal{E}_t$ based on noisy environment data.

To represent Constitutions, we employ a notation for hybrid relational domains~\cite{nitti2016probabilistic}.
Hence, $\mathcal{C}_t$ is comprised of \textit{clauses}, each being made up of a \textit{head}, an optional \textit{body}, and \textit{distribution}.
Consider the following two clauses.
\begin{align}
    \label{eq:clauses}
    p \ :: \ &r_1(a_1, \ldots, a_n) \ \text{:-} \ l_1,\ \ldots,\ l_m. \tag*{(Categorical)} \\
    &r_2(a_1, \dots, a_i) \sim f(\mathbf{\psi}) \ \text{:-} \ l_1,\ \ldots,\ l_j. \tag*{(Continuous)}
    \label{eq:distributional_clauses}
\end{align}
In the first case, the head $r_1$ is true with a probability $p$ given that all the \textit{literals} $l_k$ in the body are true.
Analogously, in the second case, head $r_2$ is distributed according to the density $f(\mathbf{\psi})$ with parameters $\mathbf{\psi}$ if its body is true.
If the right-hand side is empty, the head is regarded as a fact and distributed independently of any other symbols.
For example, let us consider the running examples from \Cref{sec:starmaps}:
\begin{align}
    \label{eq:example_clause_over}
    &0.95 \ :: \ \text{over}(\text{x}, \text{park}). \\
    &\text{distance}(\text{x}, \text{road}) \sim normal(100, 1).
    \label{eq:example_clause_distance}
\end{align}
That is, at the location x, there is a park with a probability of $0.95$, and the distance to the nearest road is about $\SI{100}{\meter}$.

\begin{figure*}[t]
    \centering
    \begin{subfigure}{0.23\linewidth}
        \includegraphics[width=\textwidth]{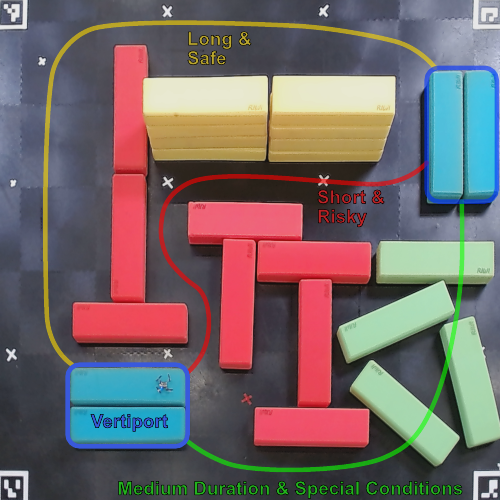}
        \caption{Annotated Testbed}
        \label{fig:testbed:annotated}
    \end{subfigure}
    \hfill
    \begin{subfigure}{0.23\linewidth}
        \includegraphics[width=\textwidth,height=\textwidth]{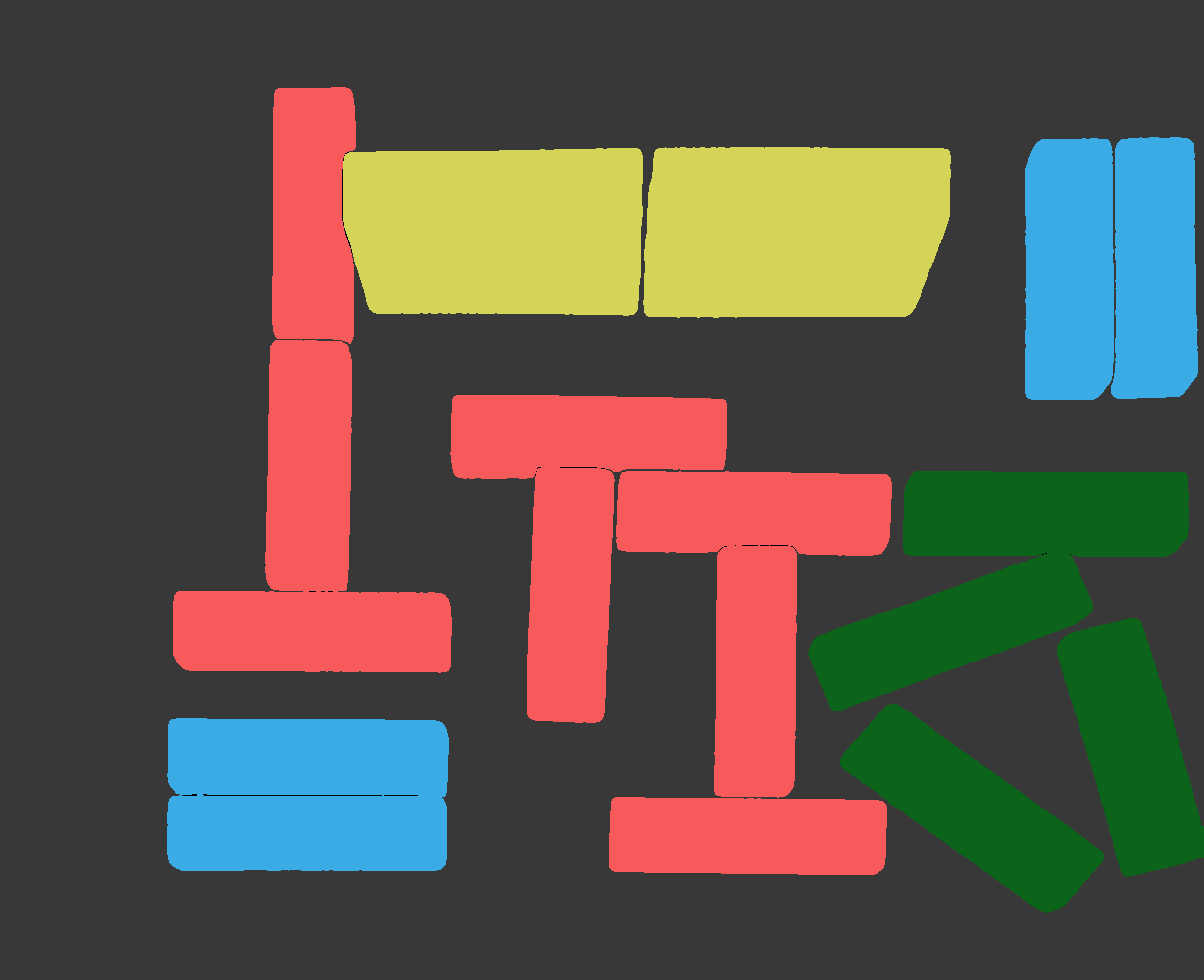}
        \caption{Segmented Testbed}
        \label{fig:testbed:segmented}
    \end{subfigure}
    \hfill
    \begin{subfigure}{0.23\linewidth}
        \includegraphics[width=\textwidth,height=\textwidth]{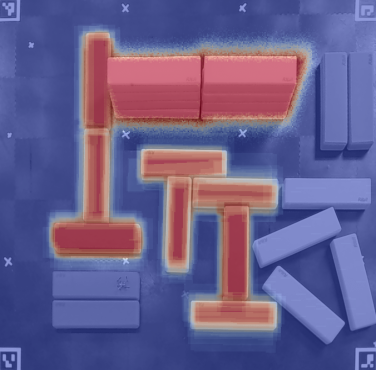}
        \caption{$P(C_t | \mathbf{x}, \mathbf{z}); v = \SI{0.5}{\meter\per\second}$}
        \label{fig:testbed:apml_slow}
    \end{subfigure}
    \hfill
    \begin{subfigure}{0.2668\linewidth}
        \includegraphics[width=\textwidth]{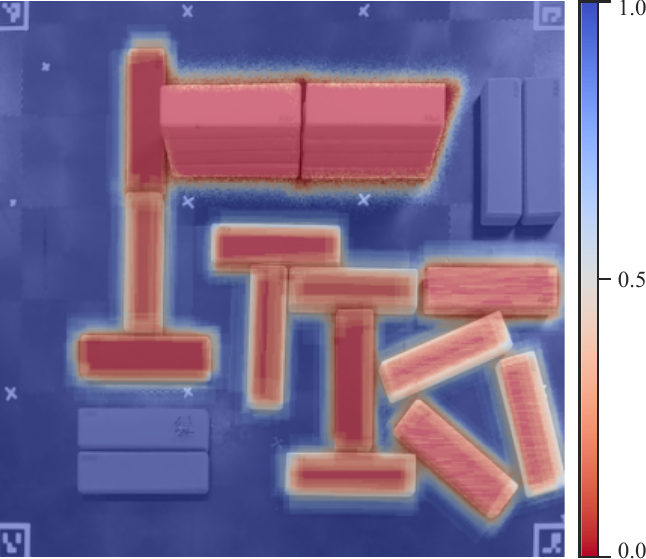}
        \caption{$P(C_t | \mathbf{x}, \mathbf{z}); v = \SI{1.0}{\meter\per\second}$}
        \label{fig:testbed:apml_fast}
    \end{subfigure}
    \caption{
        \textbf{Compliant aircraft navigation:}
        The agent may take a short, risky path in the middle, a medium-length one with speed limitations below, or a long but unconstrained path above (a).
        Via semantic segmentation (b), CoCo infers its navigation landscapes (c, d).
    }
    \label{fig:testbed}
\end{figure*}

Although one could pose arbitrary queries to this probabilistic model, we require a singular clause \textit{constitution(X, Z)} to exist, acting as a summary of the mission conditions to be applied in CoCo.
Once an off-the-shelf solver such as Prolog~\cite{prolog} or clingo~\cite{gebser2017clingo} has been applied to enumerate all solutions consisting of models $j \in \mathcal{J}$ of ground atoms $a \in \mathcal{A}$, CoCo is set up to run exact inference.

\subsection{Exact Probabilistic Inference}

In CoCo, we need to compute the probability of a pair~$(\mathbf{x}, \mathbf{z})$ to satisfy $\mathcal{C}_t$ when solving for the \textit{constitution(X, Z)} clause.
To do so, one needs to assign the probabilities $P(a = j(a) | \mathbf{x}, \mathbf{z}, \mathcal{C}_t)$ of atom $a$ to take on the value assigned by model $j$ given the state $\mathbf{x}$, measurement $\mathbf{z}$, constitution $\mathcal{C}_t$.
For CoCo, we refer to a StaR Map as laid out in Section~\ref{sec:starmaps} to provide the distribution parameters to ground atoms expressing spatial relations as in the example Equations~(\ref{eq:example_clause_over}-\ref{eq:example_clause_distance}).
Further, we assume $\mathcal{B}_t$ to be parameterized according to expert knowledge or learned separately, and for $\mathcal{P}_t$ to encode uncertainties produced by the respective perception model or sensor (see \Cref{fig:motivation} and \Cref{listing:drone_model}).
In turn, the \textbf{compliance landscape} can be computed as 
\begin{align}
    \hspace{-0.2cm}P(C_t | \mathbf{x}, \mathbf{z}) = \sum\nolimits_{j \in \mathcal{J}} \prod\nolimits_{a \in \mathcal{A}} P(a = j(a) | \mathbf{x}, \mathbf{z}, \mathcal{C}_t).
    \label{eq:wmc}
\end{align}

\subsection{Learning Doubt}

CoCo agents learn a \textbf{doubt density}, denoted $\delta_{W}(\mathbf{x} | \bm{\theta})$, which defines a conditional probability density over the state space $\mathbf{x}$, given a set of \textbf{doubt features} $\bm{\theta}$ and weights $W$.
These features encode contextual factors that influence the agent's control accuracy, such as target velocities, terrain profiles, or internal control parameters like PID gains. 
The doubt density captures motion deviations under varying operational conditions described by $\bm{\theta}$, and serves as a probabilistic representation of an agent's beliefs about its control accuracy.
We introduce two language constructs:
\begin{align}
    &\text{doubt\_feature}(c_1, \{cat_1, cat_2, \dots, cat_n\}). \tag{Categorical} \\
    &\text{doubt\_feature}(c_2, \left[a, b\right]). \tag{Continuous}
\end{align}
Hereby, the doubt features of the Constitution jointly define the interface to the neural model of $\delta_{W}(\mathbf{x} | \bm{\theta})$.
To represent $\delta_{W}(\mathbf{x} | \bm{\theta})$, we adopt the Conditional Normalizing Flows (CNF) architecture based on the Masked Autoregressive Transform~\cite{papamakarios2017masked}. 
Using CNFs enables tractable conditional density estimation, transforming a standard Gaussian into a representation of the control inaccuracies.
The conditioning on $\bm{\theta}$ enables the CNF to adjust its shape based on the mission context, allowing CoCo fine-grained modeling of state uncertainty under varying control regimes.
With state-conditioning $\{(\mathbf{x}^{(i)}, \bm{\theta}^{(i)})\}_{i=1}^N$ from recorded journeys, we fit the CNF to maximize the conditional log-likelihood
\begin{align}
    W^* = \argmax_W \sum_{i=1}^N \log \delta_{W}(\mathbf{x}^{(i)} | \bm{\theta}^{(i)}).
\end{align}

\begin{listing}[t]
    \includegraphics[width=\linewidth]{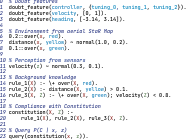}
    \caption{
        \textbf{CoCo for autonomous aircraft compliance:}
        Here, the UAS's Constitution models compliance with rules in our testbed.
    }
    \label{listing:drone_model}
\end{listing}

\subsection{Doubt Calibrated Compliance}

Building on the learned doubt density, CoCo incorporates doubt-calibrated compliance as a mechanism for aligning behavior with internal expectations and external rules. 
Here, we aim to assess how likely the agent is to comply with local operational constraints, conditioned not only on the current state but also on the context encoded by the doubt features.
To do so, we employ Monte Carlo Estimation on the doubt density $\delta_W(\mathbf{x} | \bm{\theta})$ and compliance $P(C_t | \mathbf{x}, \mathbf{z})$ such that 
\begin{align}
	P_{\bm{\theta}}(C_t | \mathbf{x}, \mathbf{z}) = \int P(C_t | \mathbf{x}, \mathbf{z}) \delta_W(\mathbf{x} | \bm{\theta}) \ d\mathbf{x}.
\end{align}
This allows to evaluate the expected compliance under varying conditions of doubt, yielding a \textbf{doubt-calibrated compliance landscape} $P_{\bm{\theta}}(C_t | \mathbf{x}, \mathbf{z})$. 
This probabilistic integration of doubt and compliance enables CoCo's principled and adaptive cost function that reflects both internal confidence and external accountability. 
The resulting policies can thus balance safety and rule adherence across a range of operating contexts.

\subsection{Constitutional Control Loop}

CoCo searches an optimal path given the doubtful compliance cost $P_{\bm{\theta}}(C_t | \mathbf{x}, \mathbf{z})$ traded off against other costs such as time or energy consumption $\mathbf{J}(\mathbf{x}) \in \mathbb{R}^J, J \in \mathbb{N}$.
That is, we obtain the optimal path $\tau^*$ such that
\begin{align}
    \tau^* = \argmin_{\tau \in \mathcal{T}} \sum_{\mathbf{x} \in \tau} \left( -\alpha \log  P_{\bm{\theta}}(C_t | \mathbf{x}, \mathbf{z}) + \bm{\beta}^T \cdot \mathbf{J}(\mathbf{x}) \right)
    \label{eq:path_optimization_cirterion}
\end{align}
where $\alpha \in \mathbb{R}$ and $\bm{\beta} \in \mathbb{R}^J$ trade off $P_{\bm{\theta}}(C_t | \mathbf{x}, \mathbf{z})$ against $\mathbf{J}(\mathbf{x})$, providing an application dependent weight to each objective.
CoCo then applies the currently chosen internal policy $\pi$ to produce motor commands $\mathbf{u}_t = \pi(\mathbf{x}_t, \mathbf{x}_t^*), \mathbf{x}_t^* \in \tau^*$.
That is, we treat $\pi$ as a black-box that is augmented via CoCo's compliance and doubt modeling techniques.

\subsection{Online Compliance Validation}
\label{sec:method:online_validation}

Not only does CoCo allow guiding an agent in a safe and compliant manner, but it also enables us to validate the agent's actions online.
To do so, we can apply the current state and measurement of the agent to the compliance landscape, resulting in the instantaneous probability of compliance.
\begin{equation}
	\hspace{-1.05em}P(C_t) = \hspace{-3pt}\iint P(C_t | \mathbf{x}_t, \mathbf{z}_t) p(\mathbf{x}_t | \mathbf{z}_{1:t-1}) p(\mathbf{z}_t | \mathbf{x}_t) \, d\mathbf{x}_t d\mathbf{z}_t
\end{equation}
This provides an additional quantity for decision-making, such as triggering emergency routines.

\begin{figure}[t]
    \centering
    \begin{subfigure}{\linewidth}
        \includegraphics[trim=1.1cm 0 0 0, width=\textwidth]{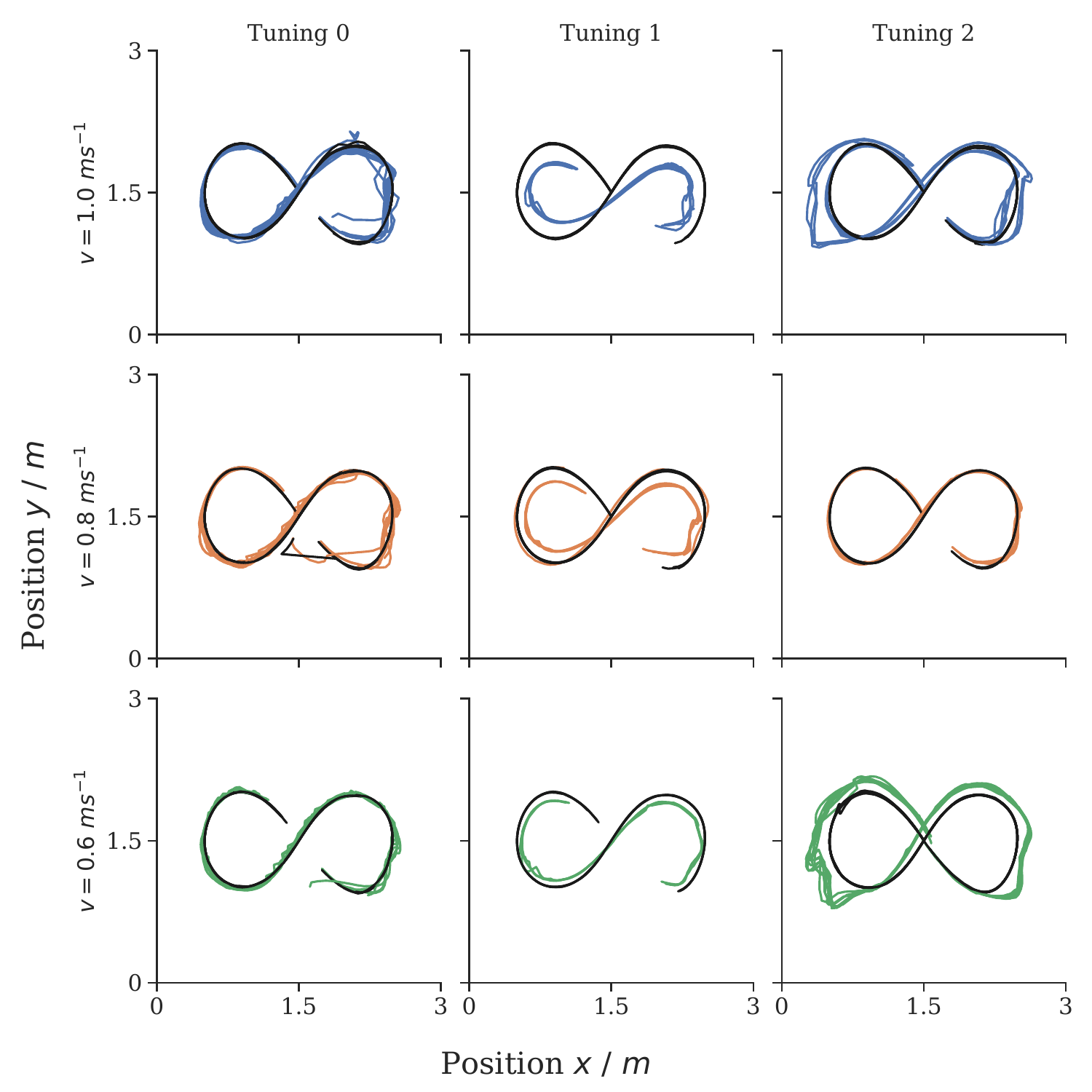}
        \caption{UAV test flights}
        \label{fig:figure_8_recordings:paths}
    \end{subfigure}\\
    \begin{subfigure}{\linewidth}
        \centering
        \includegraphics[width=\textwidth]{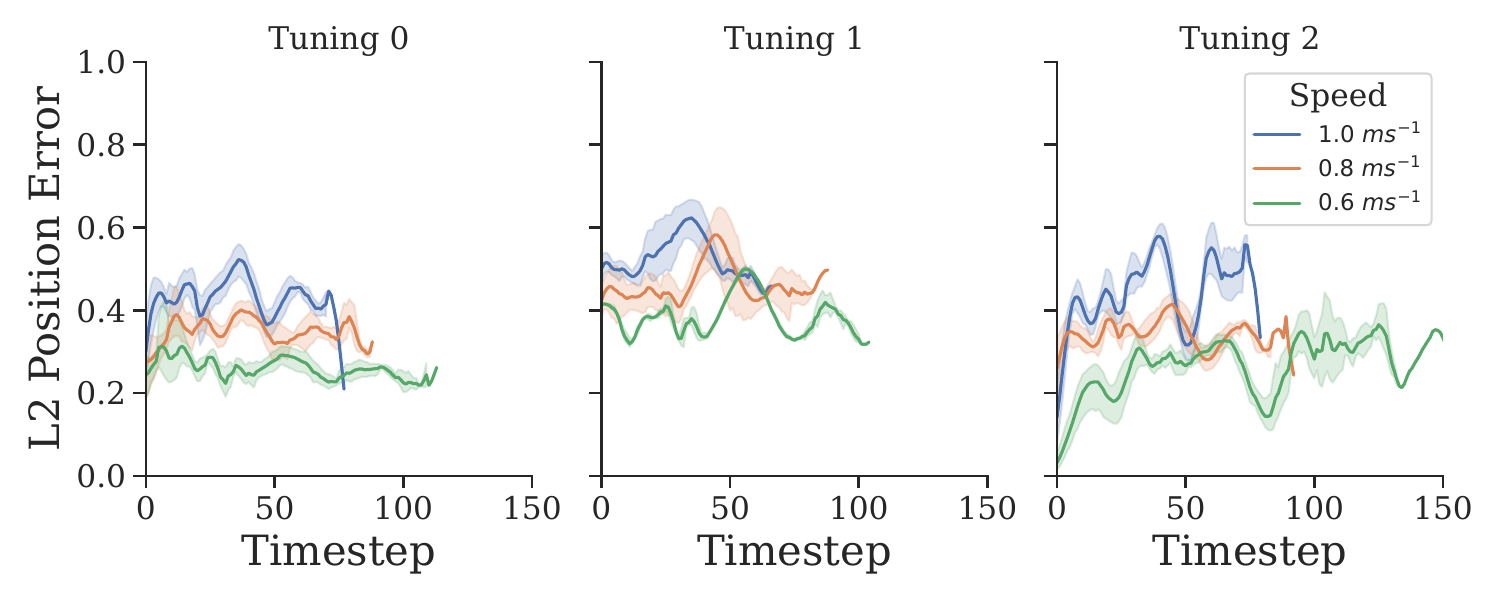}
        \caption{UAV position error over time}
        \label{fig:figure_8_recordings:error}
    \end{subfigure}
    \caption{
        \textbf{Learning from test flights:}
        CoCo learns from test flights of the UAV (a), specifically the contextualized position error (b).
    }
    \label{fig:figure_8_recordings}
\end{figure}

\begin{figure}
    \centering
    \includegraphics[width=\linewidth]{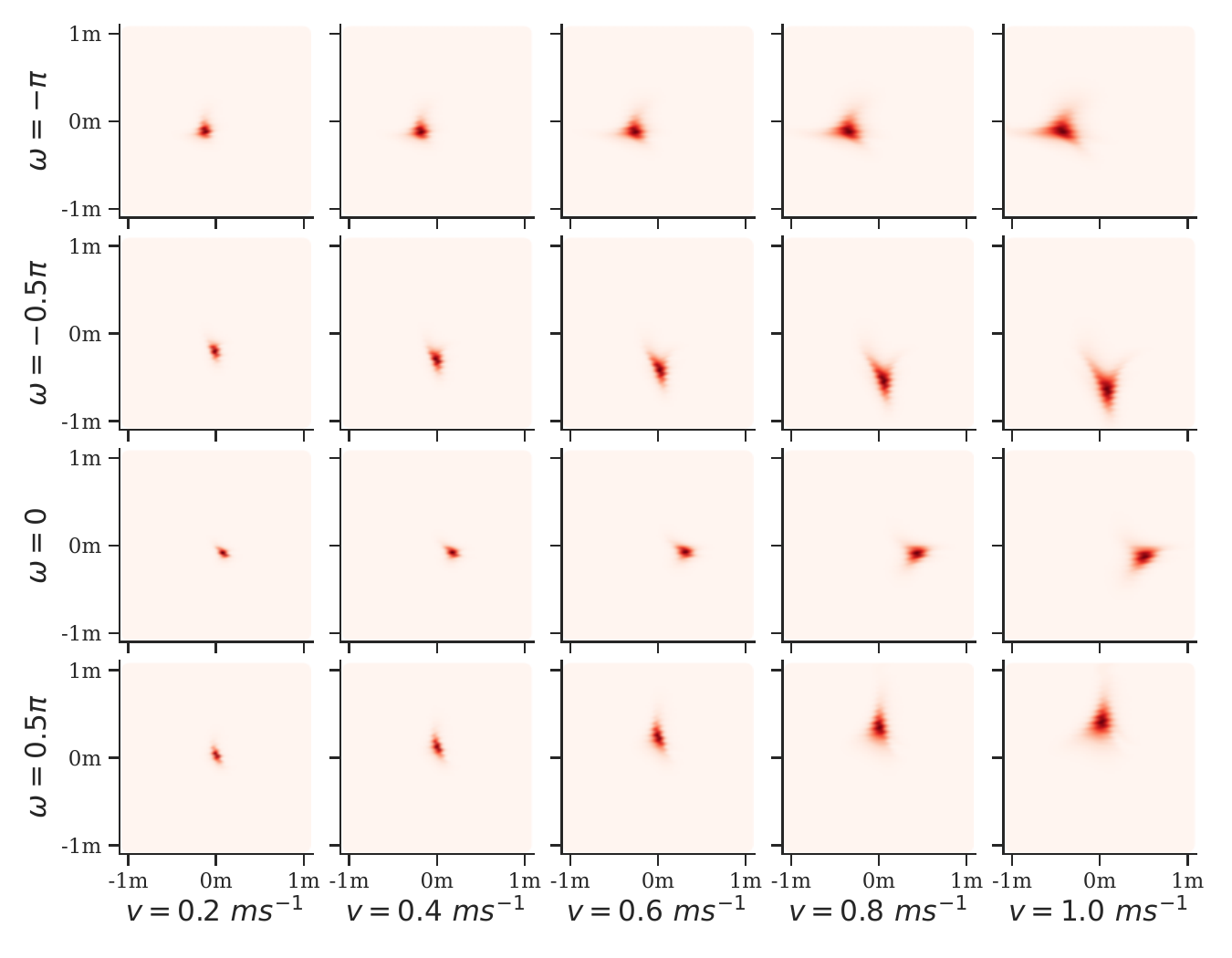}
    \caption{
        \textbf{Learning to doubt:}
        We show the doubt density fitted on test flight data as presented in \Cref{fig:figure_8_recordings} for a fixed tuning $T=0$.
    }
    \label{fig:conditional_flow}
\end{figure}

\section{Experiments}

We demonstrate our methods using a real-world drone setup, employing the Bitcraze Crazyflie 2.1~\cite{BitcrazeCrazyflie}.
\begin{enumerate}[\textbf{(Q1)}] %
    \item[\textbf{(Q1)}] Can CoCo's Constitutions capture regulations and doubt features for compliant flights?
    \item[\textbf{(Q2)}] Can CoCo learn an appropriate doubt-density from historic flight-data for safe and compliant behavior?
    \item[\textbf{(Q3)}] Does a doubt-calibrated CoCo lead to improved flight safety compared to the unconstitutional baseline?
\end{enumerate}

\begin{figure*}[t]
    \centering
    \begin{subfigure}{0.25\linewidth}
        \centering
        \includegraphics[trim=1.5cm 0 0 0, width=3.9cm]{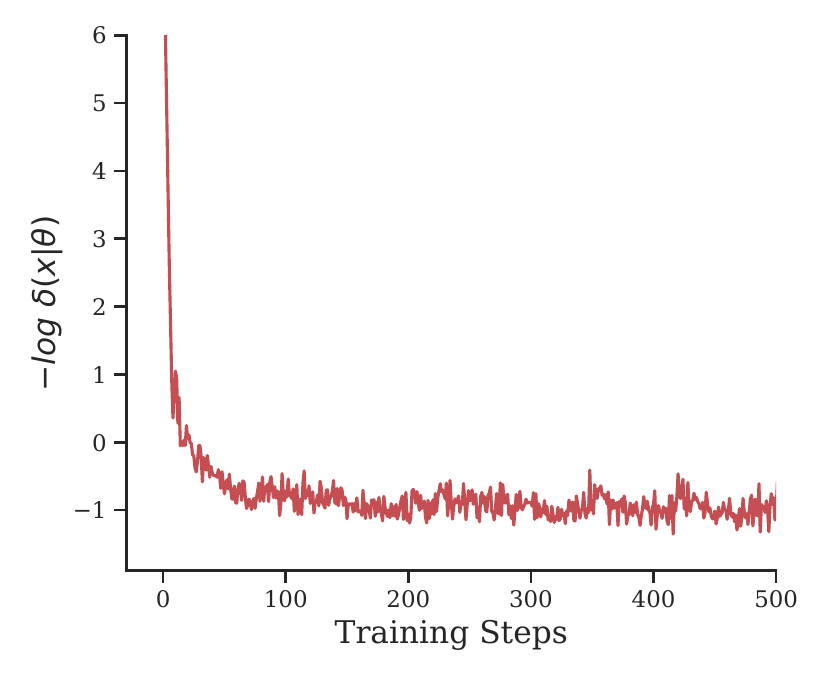}
        \caption{Loss of fitting $\delta_W(\mathbf{x} | \bm{\theta})$}
        \label{fig:doubt_expression:loss}
    \end{subfigure}
    \hfill
    \begin{subfigure}{0.23\linewidth}
        \includegraphics[trim=0 0 1.75cm 0, clip, width=\textwidth]{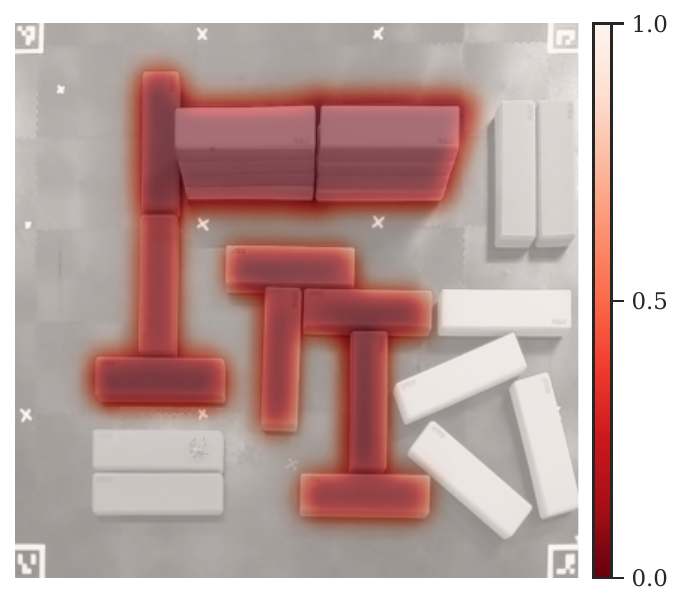}
        \caption{$P_{\bm{\theta}}(C_t | \mathbf{x}, \mathbf{z}); v = \SI{0.2}{\meter\per\second}$}
    \end{subfigure}
    \hfill
    \begin{subfigure}{0.23\linewidth}
        \includegraphics[trim=0 0 1.75cm 0, clip, width=\textwidth]{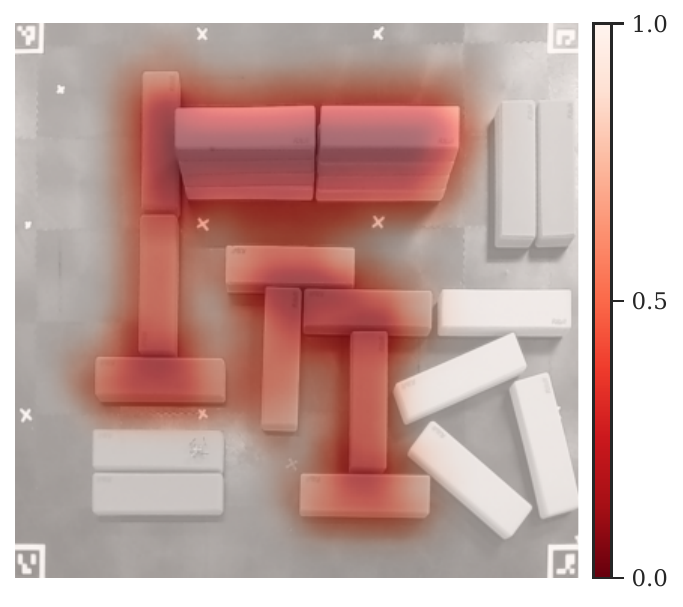}
        \caption{$P_{\bm{\theta}}(C_t | \mathbf{x}, \mathbf{z}); v = \SI{0.5}{\meter\per\second}$}
    \end{subfigure}
    \hfill
    \begin{subfigure}{0.27\linewidth}
        \includegraphics[width=\textwidth]{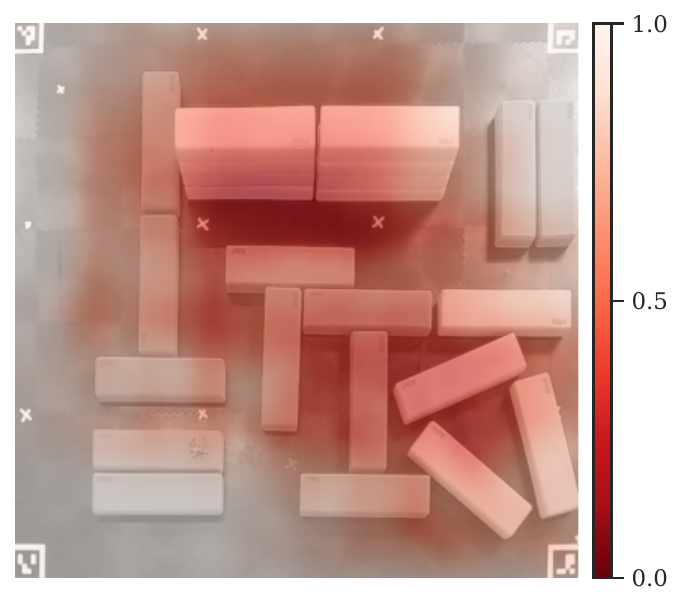}
        \caption{$P_{\bm{\theta}}(C_t | \mathbf{x}, \mathbf{z}); v = \SI{1.0}{\meter\per\second}$}
    \end{subfigure}
    \caption{
        \textbf{Doubt expression on the navigation space:}
        After fitting the Conditional Flow on the UAV's test flights (a), we show the landscapes $P_{\bm{\theta}}(C_t | \mathbf{x}, \mathbf{z})$ with the doubt density marginalized for the agent's velocity (b--d).
    }
    \label{fig:doubt_expression}
\end{figure*}

\begin{figure*}[t]
    \centering
    \begin{subfigure}{0.49\linewidth}
        \includegraphics[width=\textwidth]{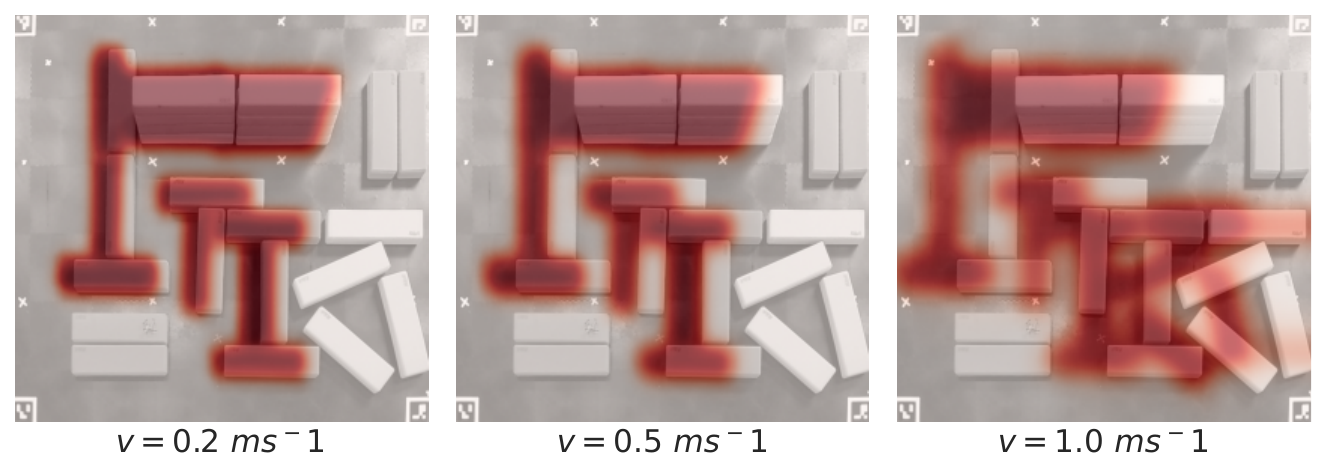}
        \caption{$P_{\bm{\theta}}(C_t | \mathbf{x}, \mathbf{z}) \text{ for } T = 0, \omega = \pi \text{, and varying } v$}
    \end{subfigure}
    \hfill
    \begin{subfigure}{0.49\linewidth}
        \includegraphics[width=\textwidth]{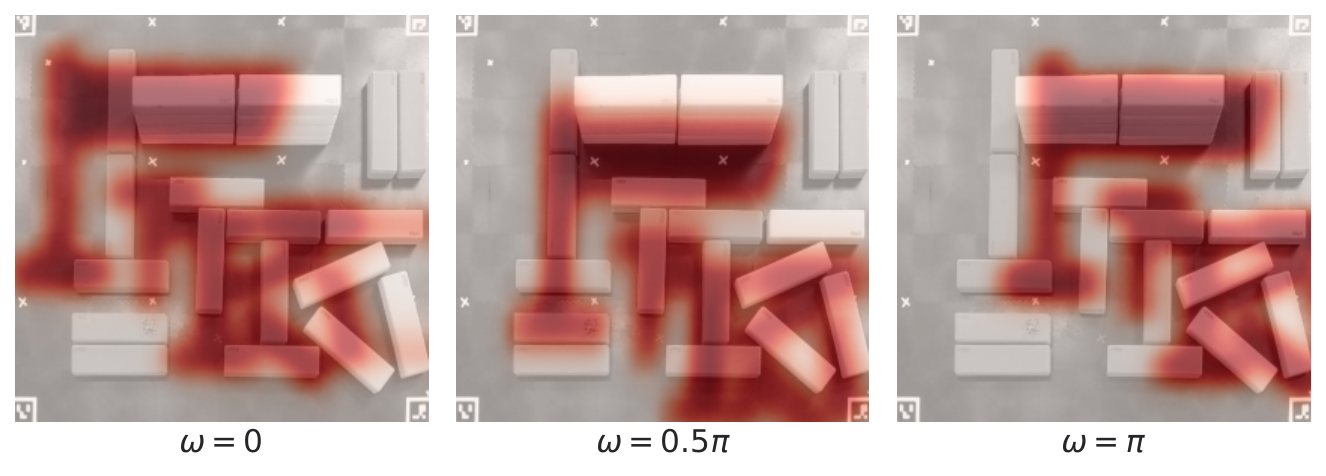}
        \caption{$P_{\bm{\theta}}(C_t | \mathbf{x}, \mathbf{z}) \text{ for } T = 0, v = \SI{1}{\meter\per\second} \text{, and varying } \omega$}
    \end{subfigure}
    \caption{
        \textbf{Affecting the compliance landscape through doubt features:}
        Transformations are induced by the doubt density and rules, e.g., speed limits, in the Constitution.
        Specifically, we see translations from varying speed (a) and rotations with varying heading angle (b).
    }
    \label{fig:doubt_augmented_navigation}
\end{figure*}

\subsection{Setup}
The experiments are conducted on the Crazyswarm2 Framework~\cite{preiss2017crazyswarm} by Bitcraze.
All experiments are performed on the Crazyflie 2.1~\cite{BitcrazeCrazyflie} nano-quadcopter using the Cascaded PID Controller~\cite{BitcrazePID} and the Extended Kalman Filter of the Crazyflie Firmware.
The doubt estimation of CoCo estimates the uncertainty over the entire control system, permitting the black-box use of the control loop as tuned by the manufacturer.
Two Valve Index Lighthouses 2.0~\cite{SteamVRBaseStations} from SteamVR are used for recording ground truth localization data.

To evaluate CoCo in a realistic scenario, we assembled a testbed as shown in \Cref{fig:testbed:annotated}.
The drone's task was to continuously commute back and forth between the two blue vertiports, starting at the bottom left, while considering the remaining three types of environment features (red, yellow, and green).
For detecting the positions and dimensions of the colored obstacles, we employed the Segment Anything Model~2 for object detection~\cite{ravi2025sam2}.
The resulting geometries shown in \Cref{fig:testbed:segmented} were then modeled by a Star Map with predetermined positional uncertainty.

\subsection{A Constitution for Compliant UAV Control}

To tackle question \textbf{Q1} and to govern the UAV's behavior in our testbed, we define a probabilistic logic program representing the agent's Constitution, as shown in \Cref{listing:drone_model}. 
This Constitution codifies high-level flight requirements and serves as a structured decision-making framework for navigating our testbed, as depicted in \Cref{fig:testbed:annotated}.

The testbed's Constitution considers three doubt features: the categorical controller tunings $T \in \{0, 1, 2\}$, the UAV’s velocity $v$, and its heading angle $\omega$.
Furthermore, it is parameterized using three main inputs:
(i) a StaR Map of the objects picked up by SAM2, providing the basic vocabulary for spatial reasoning in our Constitution,
(ii) the UAV's sensor data, here represented by its perceived velocity and updated continuously throughout the flight, and
(iii) our spatial constraints representing the traffic laws of our testbed.

Here, a compliant agent with state space $\mathbf{x} = (x, y, v, \omega)$, comprised of its 2D position, veocity $v$ and heading angle $\omega$, ought to respect no-fly zones over red blocks, needs to maintain a safety distance to yellow blocks, and slow down to a velocity $v < \SI{0.8}{\meter\per\second}$ when traveling over green blocks.
The blue blocks are designated as start and goal areas.
By integrating these constraints with the UAV’s current and desired states, the CoCo enables probabilistic inference over the state space and velocities, yielding scalar probability fields. 
These fields represent the permissibility and safety of states across the mission area under the given rules and are exemplified for varying velocities in \Cref{fig:testbed:apml_slow,fig:testbed:apml_fast}.

\begin{figure*}[t]
    \centering
    \begin{subfigure}{0.56\linewidth}
        \includegraphics[width=\linewidth]{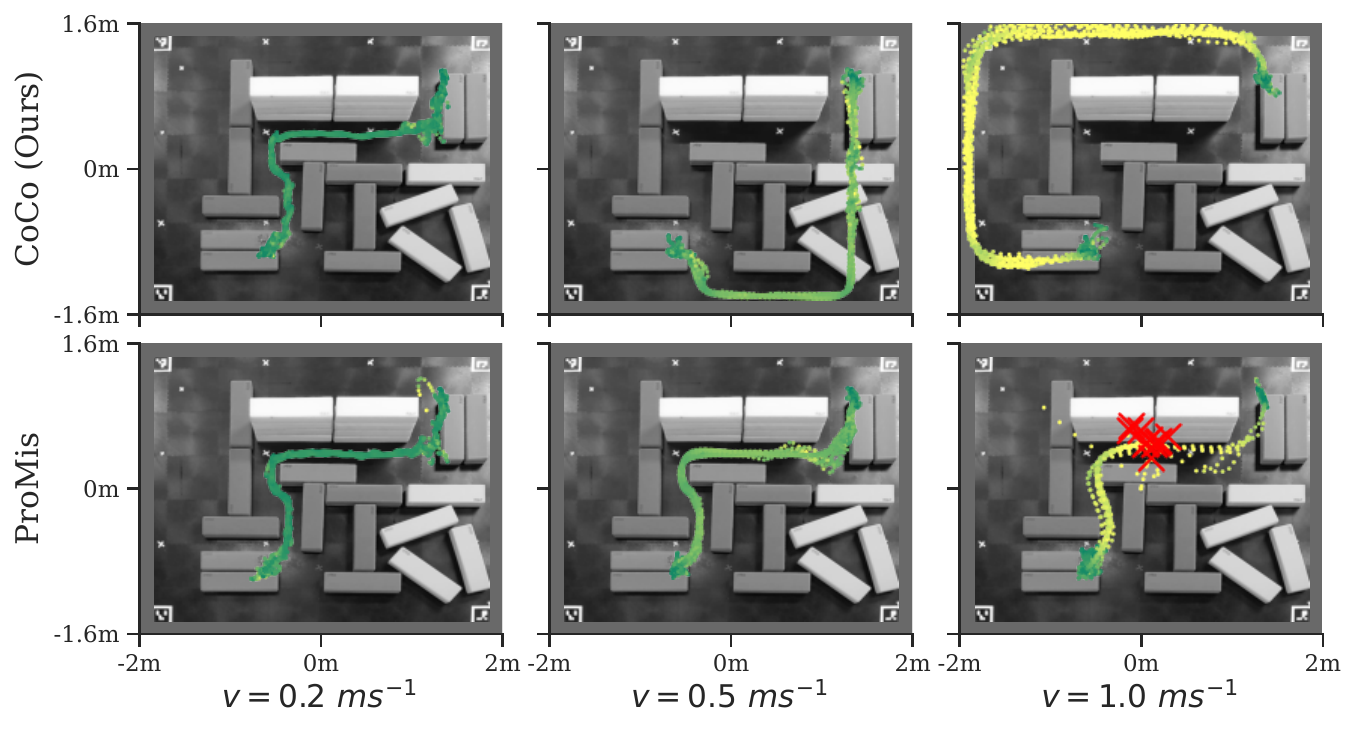}
        \caption{Constrained velocity}
        \label{fig:velocity_comparison:constrained}
    \end{subfigure}
    \hfill
    \begin{subfigure}{0.43\linewidth}
        \includegraphics[trim=0 -0.6cm 0 0, clip, width=\linewidth]{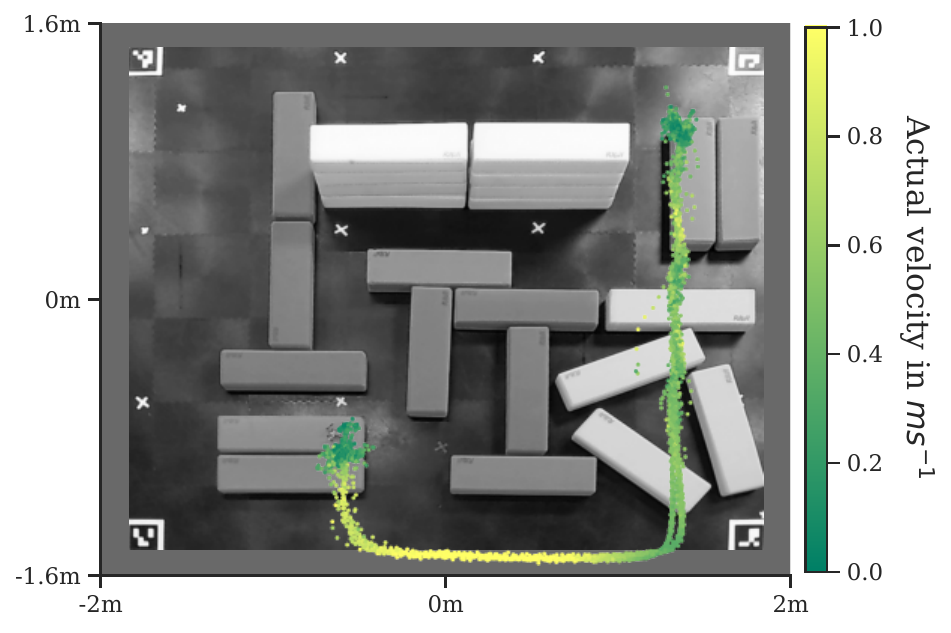}
        \caption{CoCo in control of velocity}
        \label{fig:velocity_comparison:unconstrained}
    \end{subfigure}
    \caption{
        \textbf{Constitutional Control for safe and compliant agents:}
        Because CoCo learned the UAV's capabilities, it dynamically switches paths for different target speeds. 
        In contrast, the baseline ProMis always greedily routes through the center path, crashing (red \textcolor{Maroon}{\ding{53}}) into the yellow blocks at high velocities (a).
        When CoCo takes full control, it dynamically adjusts its speed to take a compliant and safe path (b).
    }
    \label{fig:velocity_comparison}
\end{figure*}

\begin{figure*}[t]
    \centering
    \begin{subfigure}{0.56\linewidth}
        \includegraphics[width=\linewidth]{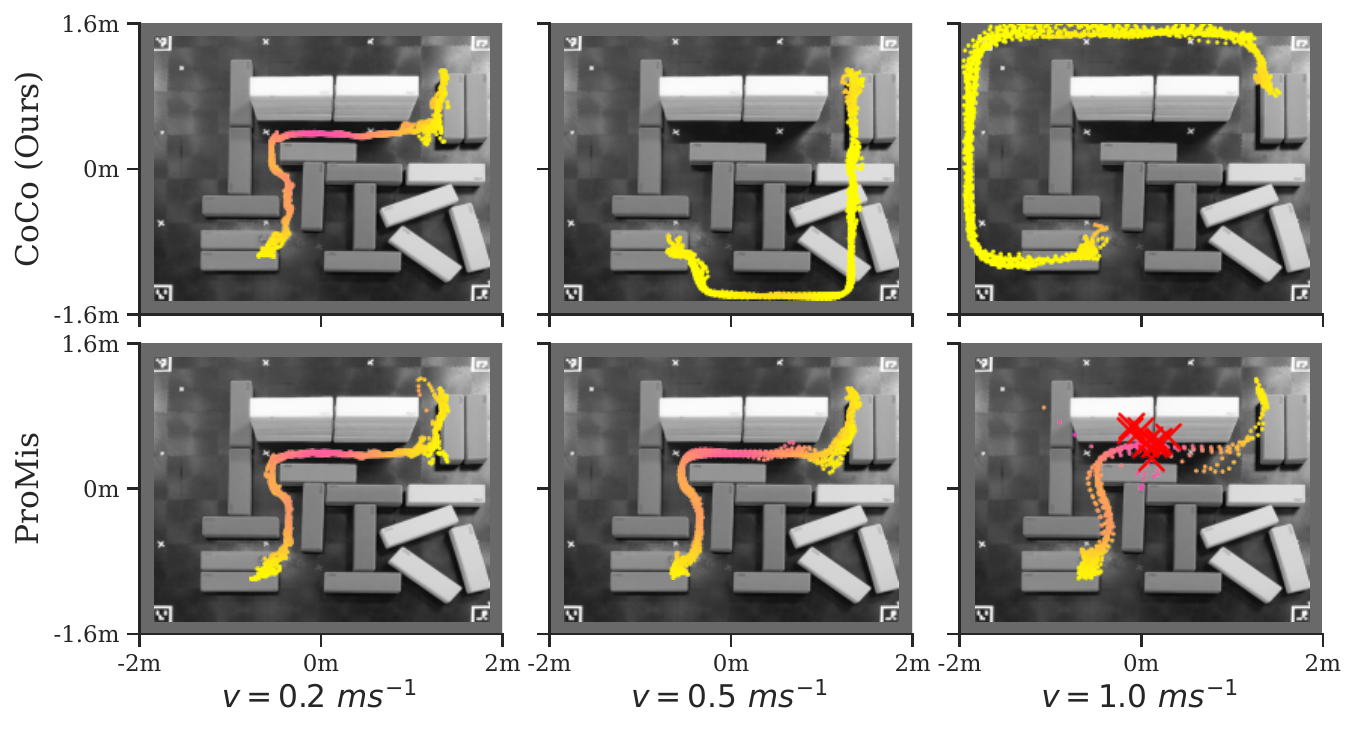}
        \caption{Constrained velocity}
        \label{fig:compliance_comparison:constrained}
    \end{subfigure}
    \hfill
    \begin{subfigure}{0.43\linewidth}
        \includegraphics[trim=0 -0.6cm 0 0, clip, width=\linewidth]{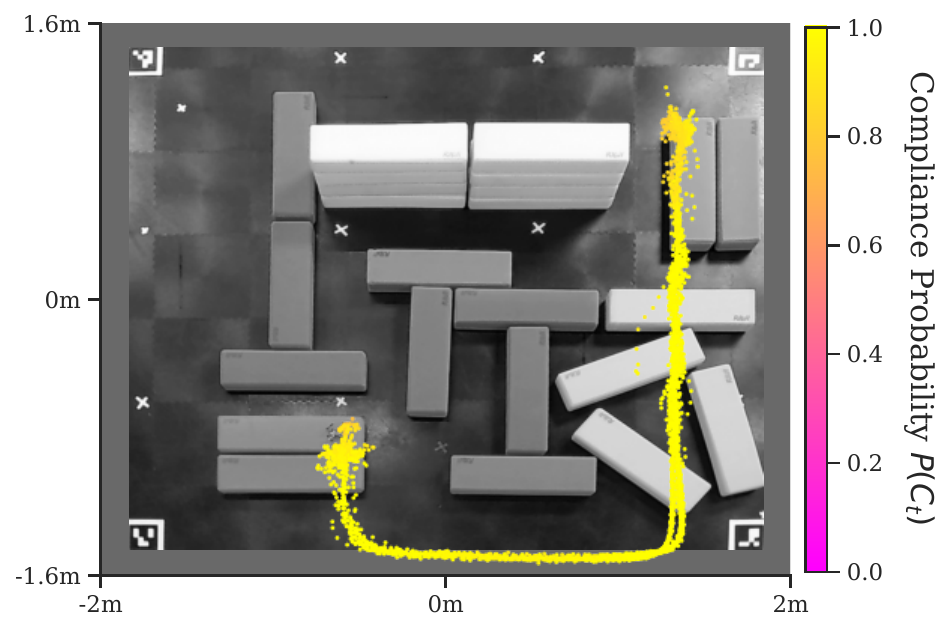}
        \caption{CoCo in control of velocity}
        \label{fig:compliance_comparison:unconstrained}
    \end{subfigure}
    \caption{
        \textbf{Compliance assurance with Constitutional Control:}
        CoCo assures a high probability of compliance compared to the baseline, both in the case of a constrained (a) and optimized (b) velocity.
    }
    \label{fig:compliance_comparison}
\end{figure*}

\subsection{Learning the Doubt Density}

We address \textbf{Q2} by conducting a series of test flights using the Crazyflie 2 platform to evaluate our approach to learning a doubt density $\delta_W(\mathbf{x} | \mathbf{\theta})$. 
Here, the doubt density captures the expected deviation between the desired and actual UAV position contextualized by a set of controller gains $T$, which are slight variations of the defaults shipped with the Crazyflie, the target velocity $v$, and heading $\omega$.
The test flights encompassed a variety of trajectories, including structured paths such as the figure-eight, which is depicted in \Cref{fig:figure_8_recordings:paths}.
For each flight, we computed the L2 norm of the position error between the desired and achieved state. 
This error metric is also visualized in \Cref{fig:figure_8_recordings:error} to illustrate representative deviations under different contexts.
Our CNF consists of 5 sequential flow layers, each composed of a Reverse Permutation followed by a Masked Autoregressive Transform with 100 hidden units.

The categorical controller tuning parameter is encoded using a one-hot representation and concatenated with the continuous features to form the full conditioning.
Training is performed using the Adam optimizer~\cite{kingma2015adam}, with the objective of maximizing the log-likelihood of the observed position errors under the modeled distribution. 
This setup allows the CNF to learn expressive representations of uncertainty in the UAV's behavior, conditioned on both control parameters and commanded trajectories.
\Cref{fig:conditional_flow} demonstrates, for controller tuning $T = 0$ and across velocities and headings, how the CoCo agent has successfully learned the doubt density, with the training loss shown in \Cref{fig:doubt_expression:loss}.

\subsection{Constitutional Control}
\label{sec:exp:control}

Regarding \textbf{Q3}, we show CoCo's improved flight characteristics based on its notion of doubt against a baseline that, while able to capture the same local traffic regulations, lacks such a mechanism.
We choose Probabilistic Mission Design (ProMis)~\cite{kohaut2023mission} as a state-of-the-art approach to integrate probabilistic logic for compliant airflight with the same laws and StaR Map relations encoded as for CoCo.
We obtain approximately optimal paths for both frameworks through A* path planning on a cost landscape comprised of travel time proportional to the agent's velocity with a weight of $\beta = \left[1\right]$ and a compliance landscape with weight $\alpha = 2$. 
While ProMis allows to model the basic landscape $P(C_t | \mathbf{x}, \mathbf{z})$, CoCo's doubt-calibration yields $P_{\bm{\theta}}(C_t | \mathbf{x}, \mathbf{z})$ as shown in \Cref{fig:doubt_expression,fig:doubt_augmented_navigation}.
We then create a navigation graph with a $300 \times 300 \times 3$ resolution to represent the 2D position of the agent and velocities $\SI{0.2}{\meter\per\second}$, $\SI{0.5}{\meter\per\second}$, and $\SI{1.0}{\meter\per\second}$.

We consider two settings to demonstrate the advantages of constitutional and doubt-calibrated control in autonomous airspaces.
First, the UAV is tasked with traveling between two vertiports at a fixed target velocity. 
While both CoCo and the baseline are able to respect the encoded traffic laws, ProMis lacks a notion of self-doubt.
Hence, at high velocities, ProMis proves unreliable and collides with the yellow blocks, prematurely ending the mission.
\Cref{fig:velocity_comparison:constrained} visualizes the trajectories along with crash locations under the baseline controller, highlighting the failure cases that occur at higher speeds.
In total, $11$ out of $45$ flights, $15$ per target velocity, of the ProMis baseline resulted in crashes ($73.33\%$ failures at high velocity).
In contrast, CoCo leverages its doubt density to reason about its own limitations. 
Hence, for each of the evaluated target velocities, CoCo selects the safest viable path that respects the Constitution's constraints, avoiding red blocks, maintaining distance from yellow ones, and considering permissible velocities above green blocks.
Hereby, CoCo avoids any crashes throughout all $45$ flights.
\Cref{fig:compliance_comparison:constrained} visualizes the compliance probability for all journeys, as defined in \Cref{sec:method:online_validation}, showing how CoCo confidently ensures the Constitution's requirements.
In the second scenario, the agent's velocity is unconstrained (\Cref{fig:velocity_comparison:unconstrained,fig:compliance_comparison:unconstrained}). 
Here, CoCo leverages the doubt model not only for path selection but also for dynamic velocity adaptation. 
The UAV begins at high speed, slows down when approaching the green blocks, and accelerates again after passing the velocity constraints, adaptively yielding safe, compliant, and time-effective missions.

\subsection{Discussion}

Applying CoCo necessitates considering the challenges of multi-objective optimization, i.e., the chosen weights of the individual cost functions will shape the agent's behavior.
Consider \Cref{fig:velocity_comparison:constrained} for $v = \SI{0.5}{\meter\per\second}$.
Comparing ProMis with CoCo, although ProMis takes on a greater risk, all flights succeed.
At such tipping points, CoCo may be convinced to take on this risk as well by decreasing $\alpha$ in \Cref{eq:path_optimization_cirterion} accordingly.
CoCo, in fact, fully subsumes ProMis in the sense that choosing $\alpha=0$ perfectly recovers the baseline behavior.
As illustrated in \Cref{fig:compliance_comparison:constrained}, CoCo allows quantifying the risks taken concerning violating the Constitution.

Hence, CoCo provides additional information for the fine-grained tuning of the behavior of the agent.
Nevertheless, since we are learning self-doubt as a context-dependent density, this tuning may be highly context-dependent, possibly requiring dedicated calibration for each new doubt feature.

\section{Conclusion}

We introduce the Constitutional Controller~(CoCo), a novel framework for autonomous agents. 
CoCo integrates neuro-symbolic reasoning in the form of deep probabilistic logic with probabilistic modeling of the agent’s own capabilities, enabling adaptable, reliable, and interpretable agents that are compliant with high-level constraints and operational norms.
We demonstrate the effectiveness of CoCo in a real-world study with Unmanned Aerial Vehicles~(UAVs). 
It outlines how logical reasoning and doubt calibration can guide decision-making processes, enabling the agent to navigate complex environments in a compliant, crash-free manner. 
The experiments show that CoCo can control its doubt features, strategically utilizing this flexibility to optimize trajectories. 

\section*{Acknowledgements}
This work received funding from Honda Research Institute EU, the Federal Ministry of Education and Research project \enquote{KompAKI} within the \enquote{The Future of Value Creation - Research on Production, Services and Work} program~(02L19C150) managed by PTKA, and the project \enquote{The Adaptive Mind} (EXC-3066/1,  No.~533717223) and Excellence Cluster EXC-3057, funded by the German Research Foundation~(DFG) under Germany's Excellence Strategy.
D.~S. Dhami received support from their Dep. of Mathematics and Computer Science and AI Systems Institute.
Map data \copyright~OpenStreetMap contributors, licensed under the ODbL.

\bibliographystyle{IEEEtran}
\bibliography{main.bib}

\end{document}